\documentclass{sig-alternate-05-2015}

\usepackage{algorithmic}
\usepackage[ruled,vlined]{algorithm2e}
\usepackage{tikz}
\usepackage{verbatim}
\usetikzlibrary{arrows,shapes}
\usepackage{multirow}
\usepackage{amsmath}
\usepackage{booktabs}
\DeclareMathOperator \real{\mathbb{R}}

\newcommand{\Ni}{({\em i})~}
\newcommand{\Nii}{({\em ii})~}
\newcommand{\Niii}{({\em iii})~}

\newcommand{\Na}{({\em a})~}
\newcommand{\Nb}{({\em b})~}

\newcommand{\ptov}{\nobreak{\sc{S2V}}}
\newcommand{\ntov}{\nobreak{\sc{N2V}}}
\newcommand{\ntovi}{\nobreak{N2V-{\sc i}}}
\newcommand{\ntovr}{\nobreak{N2V-{\sc r}}}

\newcommand{\itw}{\nobreak{IT-{\sc w}}}
\newcommand{\itu}{\nobreak{IT}}
\newcommand{\ituw}{\nobreak{IT-{\sc uw}}}

\newcommand{\reg}{\nobreak{REG}}
\newcommand{\regw}{\nobreak{REG-{\sc w}}}
\newcommand{\reguw}{\nobreak{REG-{\sc uw}}}

\newcommand{\dictregw}{\nobreak{DICTREG-{\sc w}}}
\newcommand{\dictreguw}{\nobreak{DICTREG-{\sc uw}}}

\newcommand{\ra}[1]{\renewcommand{\arraystretch}{#1}}

\begin{document}

% Copyright
\setcopyright{acmcopyright}
% DOI
%\doi{10.475/123_4}

% ISBN
%\isbn{123-4567-24-567/08/06}

%Conference
%\conferenceinfo{PLDI '13}{June 16--19, 2013, Seattle, WA, USA}

\acmPrice{\$15.00}

%
% --- Author Metadata here ---
\conferenceinfo{WWW}{'17 Perth, Western Australia.}
%\CopyrightYear{2007} % Allows default copyright year (20XX) to be over-ridden - IF NEED BE.
%\crdata{0-12345-67-8/90/01}  % Allows default copyright data (0-89791-88-6/97/05) to be over-ridden - IF NEED BE.

\title{Dis-S2V: Discourse Informed Sen2Vec}
%\subtitle{[Extended Abstract]
%}
\numberofauthors{1}  
\author{
% 1st. author
\alignauthor
Tanay Kumar Saha{$~^{\natural}$}, Shafiq Joty{$~^{\S}$}, Naeemul Hassan{$~^{\sharp}$} and Mohammad Al Hasan{$~^{\natural}$}  \\
       \affaddr{{$~^{\natural}$}Indiana University Purdue University Indianapolis, \{tksaha,alhasan@iupui.edu\} }\\
      \affaddr{{$~^{\S}$}Qatar Computing Research Institute -- HBKU, \{sjoty@qf.org.qa\} }\\
      \affaddr{{$~^{\sharp}$}University of Mississippi, \{nhassan@olemiss.edu\}}     
% 2nd. author
%\alignauthor
% Shafiq Joty\\
%        \affaddr{Qatar Computing Research Institute}\\
%        %\affaddr{P.O. Box 1212}\\
%        \affaddr{Doha, Qatar}\\
%        \email{sjoty@qf.org.qa}
% % 3rd. author
% \alignauthor 
% Naeemul Hassan\\
%        \affaddr{University of Mississippi}\\
%        %\affaddr{1 Th{\o}rv{\"a}ld Circle}\\
%        \affaddr{Oxford, Mississippi}\\
%        \email{nhassan@olemiss.edu}
% % 4th author 
% \and 
% \alignauthor 
% Mohammad Al Hasan\\
%        \affaddr{Indiana University Purdue University Indianapolis}\\
%        %\affaddr{1 Th{\o}rv{\"a}ld Circle}\\
%        \affaddr{Indiana, USA}\\
%        \email{alhasan@iupui.edu}
}

\date{24 October, 2016}
\maketitle

\begin{abstract}

Vector representation of sentences is important for many text processing tasks that involve clustering, classifying, or ranking sentences. Recently, distributed representation of sentences learned by neural models from unlabeled data has been shown to outperform the traditional bag-of-words representation. However, most of these learning methods  consider only the content of a sentence and disregard the relations among sentences in a discourse by and large.   
 
In this paper, we propose a series of novel models for learning latent representations of sentences (Sen2Vec) that consider the content of a sentence as well as inter-sentence relations. We first represent the inter-sentence relations with a language network and then use the network to induce contextual information into the content-based Sen2Vec models. Two different approaches are introduced to exploit the information in the network.  Our first approach \emph{retrofits} (already trained) Sen2Vec vectors with respect to the network in two different ways: \Ni using the adjacency relations of a node, and \Nii using a stochastic sampling method which is more flexible in sampling neighbors of a node. The second approach uses a regularizer to encode the information in the network into the existing Sen2Vec model. Experimental results show that our proposed models outperform existing methods in three fundamental information system tasks demonstrating the effectiveness of our approach. The models leverage the computational power of multi-core CPUs to achieve fine-grained computational efficiency. We make our code publicly available upon acceptance.

% \red{We have also designed a novel way of evaluating our models on a classification task, (i.e., classifying sentences into topics), and on a ranking task (i.e., extractive text summarization). If space is short, we can cut this and connect these two paragraphs.} Experimental results show that our proposed models outperform existing methods in both tasks, demonstrating the effectiveness of our approach. The models leverage the computational power of multi-core CPUs to achieve fine-grained computational efficiency. \red{We make our code publicly available for research purposes.We will upon acceptance?}
%We also design source code to be extensible, facilitate the capability to analyze the data by storing it in a database system and made our code available online. 
%Our implementation is distributed, which makes the methods quite efficient even with a network of million nodes.
%with a best absolute gain of $5.0\%$ in topic classification and $4.0\%$ in summarization. 

\end{abstract}
\vspace{-0.08in}

\begin{CCSXML}
<concept>
<concept_id>10010147.10010257.10010293.10010319</concept_id>
<concept_desc>Computing methodologies~Learning latent representations</concept_desc>
<concept_significance>500</concept_significance>
</concept>
<ccs2012>
<concept>
<concept_id>10002951.10003317.10003347.10003356</concept_id>
<concept_desc>Information systems~Clustering and classification</concept_desc>
<concept_significance>500</concept_significance>
</concept>
<concept>
<concept_id>10002951.10003317.10003347.10003357</concept_id>
<concept_desc>Information systems~Summarization</concept_desc>
<concept_significance>500</concept_significance>
</concept>

</ccs2012>
\end{CCSXML}
\ccsdesc[500]{Computing methodologies~Learning latent representations}
\ccsdesc[500]{Information systems~Clustering and classification}
\ccsdesc[500]{Information systems~Summarization}

%\ccsdesc[500]{Computer systems organization~Embedded systems}


%
%  Use this command to print the description
%
\printccsdesc

% We no longer use \terms command

\keywords{Sen2Vec; distributed representation of sentences; feature learning; discourse; retrofitting; classification; ranking; clustering}

\section{Introduction and Motivation}
\label{intro} 
% !TEX root = sen2vec-www-17.tex

%% motivate sentence representation

Many sentence-level text processing tasks rely on representing the sentences using fixed-length vectors. For example, \emph{classifying} sentences into topics using a statistical classifier like Maximum Entropy would require the sentences to be represented by vectors. Similarly, for the task of \emph{ranking} sentences based on their importance in the text using a ranking model like LexRank \cite{Erkan.Radev:04} or SVMRank \cite{Joachims:06}, one needs to first represent the sentences with fixed-length vectors. The most common method uses a bag-of-words or a bag-of-ngrams representation, where each dimension of the vector is computed by some form of term frequency statistics (e.g., \emph{tf*idf}).  

%% motivate distributed representation

Recently, \emph{distributed representations}, in the form of dense real-valued vectors, learned by neural network models from unlabeled data, has been shown to outperform the traditional bag-of-words representation \cite{Le.Mikolov:14}. Distributed representations encode the semantics of linguistic units and yield better generalization \cite{Mikolov_distributed:2013,Bengio.Ducharme:03}. However, most existing methods to devise distributed representation for sentences consider only the content of a sentence, and disregard relations between sentences in a text  by and large \cite{Le.Mikolov:14,hill-cho-korhonen:2016}. But, sentences rarely stand of their own in a well-written text. On a finer level, sentences are connected with each other by certain logical relations (e.g., \emph{elaboration}, \emph{contrast}) to express the meaning as a whole \cite{hobbs1979coherence}. On a coarser level, sentences in a text address a common topic, often covering multiple subtopics; i.e., sentences are also topically related \cite{stede2011discourse}. Our main hypothesis in this paper is that distributed representation methods for sentences should not only consider the content of the sentence but also the inter-sentence relations.  

%% Contrast with previous work
Recent work on learning distributed representations for \emph{words} has shown that semantic relations between words (e.g., synonymy, hypernymy, hyponymy) encoded in semantic lexicons like WordNet \cite{Miller:95} or Framenet \cite{Baker:98} can improve the quality of word vectors that are trained solely on unlabeled data \cite{Xu:2014,faruqui:2015:Retro,yu:dredze:2014}. Our work
in this paper is reminiscent of this line of research with a couple of crucial differences. Firstly, we are interested in representation of sentences as opposed to words, for which such resources are not readily available. Secondly, our main goal is to incorporate discourse information in the form of inter-sentence relations as opposed to semantic relations between words. These differences posit a number of new research challenges: \Ni how can we  represent inter-sentence relations? \Nii how can we effectively exploit the inter-sentence relations in our representation learning model? and \Niii how can we evaluate our model?

%% brief description of our approach
In this paper, we propose novel models for learning distributed representations for sentences that consider not only content of a sentence but relations among sentences. We represent inter-sentence relations using a network, where nodes represent sentences and edges represent similarity between the corresponding sentences. Our choice of network to represent inter-sentence relations is due to the facts that networks provide flexible ways to represent relations between any pair of sentences, and recent advances in learning distributed representations for nodes and edges in networks have shown promising results \cite{Grover.Leskovec:16,Tang.Qu:15, Tang.Qu.Mei:15,Perozzi.Al-Rfou:14}. 

We explore two different approaches to exploit the information in the network. In our first approach, we learn sentence vectors using existing content-based models, i.e., the Sen2Vec model proposed in \cite{Le.Mikolov:14}. Then we \emph{retrofit} these vectors using information encoded in network to encourage the new vectors to be similar to the vectors of related sentences and similar to their prior representations. The retrofitting is performed in two different ways: \Ni by using an efficient \emph{iterative} algorithm \cite{Talukdar.Koby.09,faruqui:2015:Retro} that incorporates adjacency relations in $1$-hop neighborhood of a node, and \Nii by training a discriminative model that seeks to preserve local neighborhoods of nodes, and in such case, the neighborhood is sampled by a flexible stochastic sampling method \cite{Grover.Leskovec:16}. In our second approach, we alter the objective function of the original Sen2Vec model with a \emph{regularizer or prior} that encourages related sentences in the network to have similar vector representations. Therefore, in this approach the vectors are learned from scratch by jointly modeling the content of the sentences and the relations between sentences.  
% how we do this
% We explore two different approaches to exploit the information in the graph. In our first approach, we first learn sentence vectors using existing content-based models, i.e., the Sen2Vec model proposed in \cite{Le.Mikolov:14}. Then we \emph{retrofit} these vectors on the graph to encourage the new vectors to be similar to the vectors of related sentences and similar to their prior representations. The retrofitting is performed in two different ways: \Ni by using an efficient \emph{iterative} algorithm \cite{Talukdar.Koby.09,faruqui:2015:Retro} that incorporates adjacency relations between nodes, and \Nii by training a discriminative model that seeks to preserve local neighborhoods of nodes, where the neighborhood is sampled by a flexible stochastic sampling method \cite{Grover.Leskovec:16}. 

% In our second approach, we alter the objective function of the original Sen2Vec model with a \emph{regularizer or prior} that encourages related sentences in the graph to have similar vector representations. Therefore, in this approach the vectors are learned from scratch by jointly modeling the content of the sentences and the relations between sentences.  

% evaluation
Different approaches to evaluate sentence representation methods have been proposed including sentence-level prediction tasks (e.g., sentiment classification, paraphrase identification) and sentence-pair similarity computation task \cite{hill-cho-korhonen:2016,Le.Mikolov:14}. Since existing representation methods encode vectors for each sentence independently, these evaluation methodologies are trivial. In contrast, our learning methods exploit inter-sentence similarities, which can be constrained by document boundaries. Therefore, we require datasets containing documents with sentence-level annotations.  

%Therefore, for supervised prediction tasks, we require training and test datasets containing documents with sentence-level annotations, and for unsupervised tasks, we require annotations only for the test datasets in order to be able to evaluate our models. 

We evaluate our models on three different types of tasks: \emph{classification, clustering} and \emph{ranking}.  In particular, we consider the tasks of classifying and clustering sentences into topics, and of ranking sentences in a document to create an extractive summary of the document (i.e., by selecting the top-ranked sentences). %While classification is performed with supervised models, clustering and ranking are done with unsupervised models. 
There are standard datasets with document-level topic annotations (e.g., \emph{Reuters-21578}, \emph{20 Newsgroups}). However, to our knowledge, no dataset exists with topic annotations at the sentence level. We generate sentence-level topic annotations from the document-level ones by selecting subsets of sentences that can be considered as representatives of the document and label them with the same document-level topic label. We use the standard DUC 2001 and 2002 datasets to evaluate our models on the summarization task, where we compare the system-generated summaries with the human-authored summaries.          

%\red{Our results show that ... Add contribution}\\
Our experimental results on these tasks demonstrate that the models which induce information encoded in the discourse network in the form of inter-sentence relations during learning (i.e., the regularized models) consistently outperform the content-only baseline by a good margin in all tasks, whereas the retrofitted models perform comparatively better on the \emph{classification} and on the \emph{clustering} tasks.

%The models which incorporate network information during post-processing performs better than content only model in \emph{classification} and \emph{clustering} tasks. 

\section{Methodology}
\label{methods} 
%\red{
%- Primary: Shafiq (writing); Tanay (coding)  \\
%- Assist: Naffi
%}

Let $G=(V, E, W)$ be a discourse graph, where a node $\mathbf{v} \in V$ represents a sentence and the edge weight $w(\mathbf{u}, \mathbf{v}) \in W$ reflects some form of similarity between sentences $\mathbf{u}$ and $\mathbf{v}$. A sentence $\mathbf{v}$ is a sequence of words $(v_1, v_2 \cdots v_m)$, each coming from a vocabulary $\Omega$, i.e., $v_i \in \Omega$. We define $N(\mathbf{v})$ as the set of neighbors of a sentence $\mathbf{v}$ in $G$. Let $\phi:V \rightarrow \real^{d}$ be the mapping function from sentences to their distributed representations, i.e., real-valued vectors of $d$ dimensions. Our goal is to learn $\phi$ by exploiting information from two different sources: \Ni the content of the sentence, $\mathbf{v} = (v_1, v_2 \cdots v_m)$; and \Nii the neighborhood of the sentence, $N(\mathbf{v})$.   

In the following, we first describe how we construct the discourse graph $G$ for our problem (Section \ref{sec:dis_graph}). We then briefly describe existing sentence representation methods that consider only the sentence contents (Section \ref{sec:sen2vec}). In Section \ref{sec:node2vec}, we demonstrate a network-based representation method that considers only the neighborhood information. Finally, we present our retrofitting (Section \ref{sec:retro_sen2vec}) and regularized methods (Section \ref{sec:reg_sen2vec}) that incorporate both content and network neighborhood information into a single model. 

\subsection{Discourse Graph Formulation} \label{sec:dis_graph}

Let $D = \{\mathbf{v}_1, \mathbf{v}_2, \cdots, \mathbf{v}_n \}$ be the set of sentences in our dataset, which constitutes the nodes in the discourse graph. The edge weights $w(\mathbf{v}_i, \mathbf{v}_j)$ are computed by measuring the similarity  between the corresponding sentences, $\sigma(\mathbf{v}_i, \mathbf{v}_j)$.  Here $\sigma$ denotes a similarity metric (e.g., \emph{Cosine}, \emph{Jaccard}). 

For constructing the network, we distinguish between two types of edges: \Ni \emph{intra-document edges}, i.e., edges between sentences of the same document, and \Nii \emph{across-document edges}, i.e., edges between sentences of different documents. Different thresholding parameters can be set depending on whether the edges are intra--document or across-documents.

%Our analysis is general and applies to any (un)directed,
%(un)weighted network. 

%\subsection{Problem Definition} \label{sec:problem}

\subsection{Content-based Model (S2V)} \label{sec:sen2vec}
\begin{figure}[t!]
\centering
\includegraphics[width=\columnwidth]{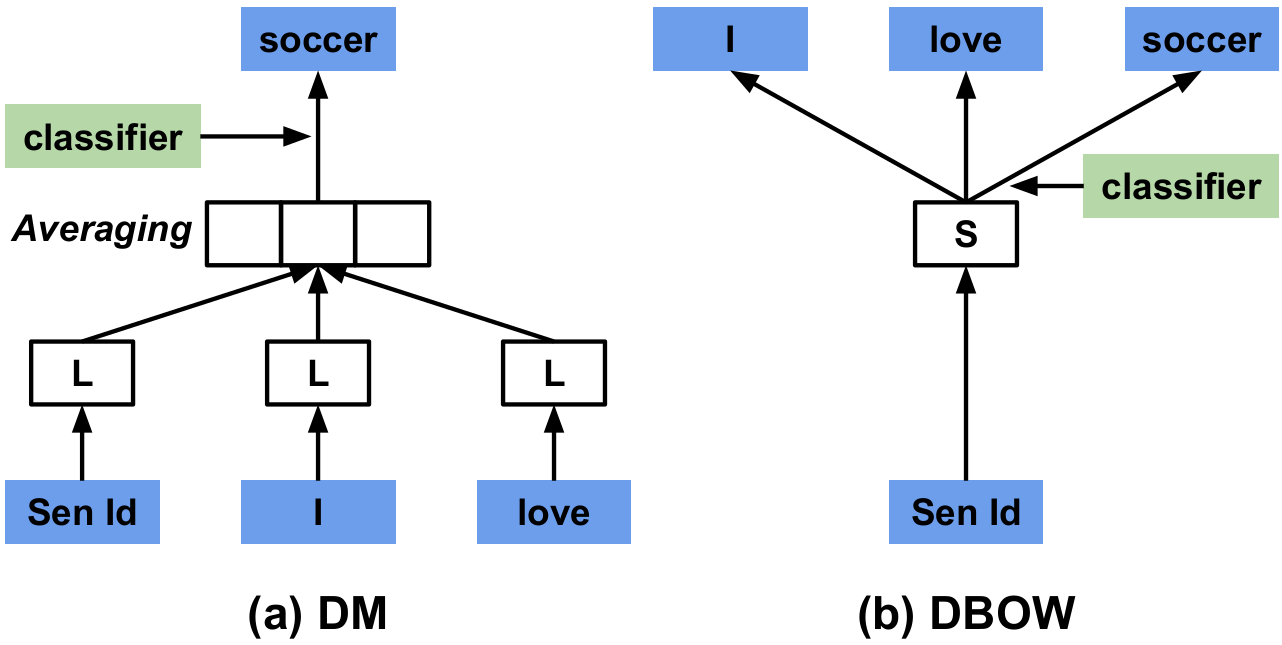}
\vspace{-0.2in}
\caption{Distributed memory (DM) and Distributed bag of words (DBOW) versions of Sen2Vec.}
%\vspace{-0.15in}
%\caption{Distributed Memory (DM) and Distributed Bag-of-words (DBOW) version of Sen2Vec model. For DM, the task while learning representation is to predict the word ``soccer'' given ``Sen Id (sentence id)'', ``I'' and ``love''; whereas, in DBOW, the model needs to predict all the words: ``I'', ``love'', ``soccer'' given the sentence id.}
\label{s2v}
\end{figure}

We use the Sen2Vec model proposed in \cite{Le.Mikolov:14} as our baseline model, which is trained solely based on the contents of the sentences. This approach concatenates the vectors learned by the two models shown in Figure \ref{s2v}: \Na a distributed memory (DM) model, and \Nb a distributed bag of words (DBOW) model. In the DM model, every sentence in the dataset $D$ is represented by a $d$ dimensional vector in a shared lookup matrix $S \in \real^{n \times d}$. Similarly, every word in the vocabulary $\Omega$ is represented by a $d$ dimensional vector in another shared lookup matrix $L \in \real^{|\Omega| \times d}$. Given an input sentence $\mathbf{v} = (v_1, v_2 \cdots v_m)$, the corresponding sentence vector from $S$ and the corresponding word vectors from $L$ are averaged to predict the next word in a context. More formally, let $\phi$ denote the mapping from sentence and word ids to their respective vectors in $S$ and $L$, the DM model minimizes the following objective (negative log likelihood):

\vspace{-0.15in}
\begin{eqnarray}
J(\phi) &=& - \sum_{t=k}^{m-k} \log P(v_t|\mathbf{v}; v_{t-k+1},\cdots,v_{t-1}) \label{eq:dm}\\
        &=& - \sum_{t=k}^{m-k} \log \frac{\exp (\omega(v_t)^T \mathbf{z})} {\sum_{i} \exp (\omega(v_i)^T \mathbf{z})} 
\end{eqnarray}

\noindent where $\mathbf{z}$ is the average of $\phi(\mathbf{v}), \phi(v_{t-k+1}),\cdots, \phi(v_{t-1})$ \emph{input} vectors, and $\omega(v_t)$ is the \emph{output} vector representation of word $v_t \in \Omega$. The sentence vector $\phi(\mathbf{v})$ is shared across all (sliding window) contexts extracted from the same sentence, thus acts as a distributed memory.

In stead of predicting the next word in the context, the DBOW model predicts the words in the context independently given the sentence id as input. More formally, DBOW minimizes the following negative log likelihood objective:  

\vspace{-0.15in}
\begin{eqnarray}
J(\phi) &=& - \sum_{t=k}^{m-k} \sum_{j=t-k+1}^{t} \log P(v_j|\mathbf{v}) \label{eq:dbow}\\
        &=& - \sum_{t=k}^{m-k} \sum_{j=t-k+1}^{t} \log \frac{\exp (\omega(v_j)^T \phi(\mathbf{v}))} {\sum_{i} \exp (\omega(v_i)^T \phi(\mathbf{v}))} 
\end{eqnarray}

\subsection{Network-based Model (N2V)} \label{sec:node2vec}
The network-based representation model considers only the neighborhood information of the nodes. We use the recently proposed Node2Vec method \cite{Grover.Leskovec:16}. It uses  the skip-gram model of Word2Vec \cite{Mikolov.Chen:13} with the intuition that nodes within a graph \emph{context (or neighborhood)} should have similar representations.  The negative log likelihood objective of the skip-gram model for graphs can be defined as:

% propose a representation method called Node2Vec for nodes in a graph. Node2Vec 

%However, as opposed to Word2Vec, the skip-gram model in Node2Vec uses a symmetric loss function (i.e., no distinction between input and output embeddings) to encourage nodes within a graph \emph{context} to have similar representations. 

\vspace{-0.15in}
\begin{eqnarray}
J(\phi) &=& - \sum_{\mathbf{v} \in V} \log P(N(\mathbf{v})|\phi(\mathbf{v})) \\
        &=& - \sum_{\mathbf{v} \in V} \sum_{\mathbf{n}_i \in N(\mathbf{v})} \log P(\mathbf{n}_i|\phi(\mathbf{v}))  \label{eq:n2v_int} \\
%        &=& \sum_{u \in V} \sum_{n_i \in N(u)} \log P(n_i|\phi(u))  \nonumber \\
        &=& - \sum_{\mathbf{v} \in V} \sum_{\mathbf{n}_i \in N(\mathbf{v})} \log \frac{\exp(\omega(\mathbf{n}_i)^T \phi(\mathbf{v}))}{\sum_{\mathbf{x} \in V} \exp(\omega(\mathbf{x})^T \phi(\mathbf{v}))} \label{eq:n2v}
\end{eqnarray}

\noindent where as before, $\phi$ and $\omega$ denote the \emph{input} and the \emph{output} vector representations of the nodes (sentences). The neighboring nodes $N(\mathbf{v})$ forms the \emph{context} for node $\mathbf{v}$. Node2Vec uses a biased random walk which adaptively combines breadth first search (BFS) and depth first search (DFS) to find the neighborhood of a node. The walk attempts to capture two properties of a graph often used for prediction tasks in networks: \Ni homophily and \Nii structural equivalence. According to homophily, nodes in the same group or community should have similar representations (e.g., sentences in a topic). Structural equivalence suggests that nodes with similar structural roles (hub, bridge) should have similar representations (e.g., central sentences). In a real-world network, nodes exhibit mixed properties.

%the mapping from node (i.e., sentence) to its \emph{input} vector representation, and $\omega$ denotes the \emph{output} vector.  

%and $\omega$ are the input and output vector representations, respectively. 

%BFS samples immediate neighbors that can capture structural equivalence in a graph. However, given a sample size, it explores only a small portion of the graph. DFS, on the other hand, samples neighbors with increasing distances, thus explores the homophily aspect of the network with more depth. 

\tikzstyle{vertex}=[circle,fill=black!15,minimum size=20pt,inner sep=0pt]
\tikzstyle{selected_vertex} = [vertex, fill=blue!25]
\tikzstyle{selected_vertex2} = [vertex, fill=green!30]
\tikzstyle{edge} = [draw,thick,-]
\tikzstyle{weight} = [font=\small]

\begin{figure}[tb!]
\centering
\begin{tikzpicture}[scale=1.5, swap]
    % First we draw the vertices
%    \foreach \pos/\name in {{(3,1)/g}, {(5,1)/c},
%                            {(4,0)/b}, {(3,-1)/e}, {(5,-1)/d}}
%        \node[vertex] (\name) at \pos {$\name$};
	 \node[selected_vertex] (b) at (4,0) {$\mathbf{u}$};
     \node[selected_vertex2] (g) at (3,1) {$\mathbf{p}$};
	 \node[vertex] (c) at (5,1) {$\mathbf{s}$};
	 \node[vertex] (d) at (5,-1) {$\mathbf{t}$};
     \node[vertex] (e) at (3,-1) {$\mathbf{w}$};
% Connect vertices with edges and draw weights
%    \foreach \source/ \dest /\weight in {b/g/$\frac{1}{p}$, b/c/5,
%                                         e/b/8, d/b/9}
%        \path[edge] (\source) -- node[weight] {$\weight$} (\dest);
        \path[edge] (g) -- (c);
        \path[edge] (b) -- node[weight] {$\mu = \frac{1}{r}$} (g);
        \path[edge] (b) -- node[weight] {$\mu = 1$} (c);
        \path[edge] (b) -- node[weight] {$\mu = \frac{1}{f}$} (e);
        \path[edge] (b) -- node[weight] {$\mu = \frac{1}{f}$} (d);
\end{tikzpicture}
\caption{Random walk for generating samples.}
\vspace{-0.1in}
\label{fig:rw}
\end{figure}
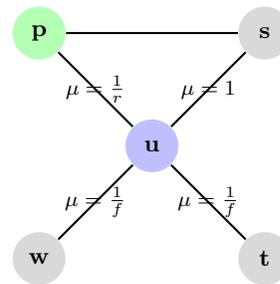
  
The controlled random walk in Node2Vec interpolates between BFS and DFS to generate \emph{multiple} samples of $N(\mathbf{v})$ for each source node $\mathbf{v}$. Let $\mathbf{u}$ be a node just traversed from $\mathbf{p}$ (Figure \ref{fig:rw}) in a random walk that started from a source node $\mathbf{v}$. The walk is controlled by (unnormalized) transition probabilities $\tau_{u,x} = \mu_{r,f}(\mathbf{p},\mathbf{x}).w_{v,x}$, where   

\vspace{-0.05in}
\begin{equation}
    \mu_{r,f}(\mathbf{p},\mathbf{x}) =
    \begin{cases}
      \frac{1}{r}, & \text{if}\ \delta_{p,x} = 0 \\
      1, & \text{if}\ \delta_{p,x} = 1 \\
      \frac{1}{f} & \text{if}\ \delta_{p,x} = 2 \\
    \end{cases}
    \label{eq:rw_cases}
\end{equation}

\noindent where  $\delta_{p,x}$ is the distance from node $\mathbf{p}$ to $\mathbf{x}$. The return parameter $r$ controls the likelihood of revisiting a node; large value will reduce duplicate samples and small value will keep the search local. The forward parameter $f$ controls the distance covered by the walk; for $f>1$, the walk will act as BFS and for $f<1$, it will act as DFS. Given a graph, we can precompute $\tau$ by keeping three possible values corresponding to Equation \ref{eq:rw_cases} for each edge transition. 

We can apply Node2Vec directly to the discourse graph to learn vector representations for the sentences. This adaptation of the model considers only \emph{inter-sentence relations} defined by the graph. Since the graph in our case is formed by matching the contents of the sentences, this method indirectly takes \emph{contents} of individual sentences into account. 

In order to incorporate contents directly, we could have a variant of the Node2Vec model, where in stead of randomly initializing the vectors, we initialize them with the precomputed vectors from the content-based model described in Section \ref{sec:sen2vec}. However, we hypothesize that a more effective approach is to consider both sources of information simultaneously in a controllable joint framework.   

%Sen2Vec method of \cite{Mikolov.Chen:13}, which is a purely content-based model. 

%%%% ENO of NODE2VEC ######################################################################

\subsection{Dis-S2V by Retrofitting} \label{sec:retro_sen2vec}

We explore the general idea of \emph{retrofitting} \cite{faruqui:2015:Retro} to incorporate information from both the content and the neighborhood of a node (or a sentence) in a joint learning framework. Let $\phi ^ \prime(\mathbf{v})$ denote the vector representation for sentence $\mathbf{v}$ that has already been learned by our content-based method in Section \ref{sec:sen2vec}. Our aim is to retrofit this vector on the discourse graph $G$ such that the revised vector $\phi(\mathbf{v})$: \Ni is also similar to the prior vector $\phi ^ \prime(\mathbf{v})$, and \Nii is similar to the vectors of its adjacent nodes $\phi(\mathbf{u})$. To this end, we define the following objective function to minimize:

%\small
\begin{algorithm}[t]
\SetAlgoLined
\SetKwInOut{Input}{Input}
\SetKwInOut{Output}{Output}
\Input{\newline - Graph $G=(V, E, W)$
       \newline - Prior vectors $\phi ^ \prime$
       \newline - Probabilities $\alpha_v$ and $\beta_v$ for all $\mathbf{v} \in V$
       }
\Output{Retrofitted vectors $\phi$}
       
\hspace{-0.4cm} 1. $\phi \leftarrow \phi ^ \prime$  \tcp{initialization}
\hspace{-0.4cm} 2. \Repeat {convergence}{
\hspace{-0.4cm} ~~\For {all $\mathbf{v} \in V$} { 
%  $\hat{\mathbf{X}}_v \leftarrow  \frac{\alpha_v \mathbf{X}_v + \sum_{(u,v) \in E} \left( \beta_v W_{u,v} + \beta_u W_{v,u} \right) \hat{\mathbf{X}}_u}{\alpha_v + \sum_{(u,v) \in E} \left( \beta_v W_{u,v} + \beta_u W_{v,u} \right) }$ \\ 
   $\phi(\mathbf{v}) \leftarrow  \frac{\alpha_v \phi ^ \prime(\mathbf{v}) + \sum_{u} \left( \beta_v W_{v,u} + \beta_u W_{u,v} \right) \phi(\mathbf{u})}{\alpha_v + \sum_{u} \left( \beta_v W_{v,u} + \beta_u W_{u,v} \right)}$ \\
 }
} % end of repeat 
\caption{Jacobi method for retrofitting.}
\label{alg:ret}
\end{algorithm}
%\normalsize

\begin{equation}
J(\phi) = \sum_{\mathbf{v} \in V} \alpha_v ||\phi(\mathbf{v}) - \phi ^ \prime(\mathbf{v})||^2 + \hspace{-0mm} \sum_{(u,v)\in E} \hspace{-0mm} \beta_v W_{u,v} ||\phi(\mathbf{u}) - \phi(\mathbf{v})||^2 \label{eq:ret}
%&=& \sum_{u \in V} \alpha_u ||\mathbf{X_u} - \mathbf{\hat{X}_u}||^2 +  \sum_{(u,v)\in E} W'_{u,v} ||\mathbf{\hat{X}_u} - \mathbf{\hat{X}_v}||^2 \label{eq:ret} 
%&=& \sum_d \big[ (X_d - \hat{X}_d)^T A (X_d - \hat{X}_d) +  \hat{X}_d^T B \hat{X}_d \big]  
\end{equation} 

\noindent where $\alpha$ values control the strength to which the algorithm should match the prior vectors, and $\beta$ values control the degree of smoothness based on the graph similarity. The quadratic cost in Equation \ref{eq:ret} is convex in $\phi$, and has a closed form solution \cite{Talukdar.Koby.09}. The closed form expression requires an inversion operation, which could be expensive for big graphs. A more efficient way is to use the  Jacobi method, an online algorithm to solve the Equation iteratively. Algorithm \ref{alg:ret} gives a pseudocode, which has the following update:

\begin{equation}
 \phi(\mathbf{v}) \leftarrow  \frac{\alpha_v \phi ^ \prime(\mathbf{v}) + \sum_{u} \left( \beta_v W_{v,u} + \beta_u W_{u,v} \right) \phi(\mathbf{u})}{\alpha_v + \sum_{u} \left( \beta_v W_{v,u} + \beta_u W_{u,v} \right)} \label{eq:ret2} 
\end{equation}

\noindent If for all $\mathbf{v}$$\in$$V$, $\alpha_v$$=$$\alpha$ and $\beta_v$$=$$\beta$, Equation \ref{eq:ret2} simplifies to:\footnote{Under this condition, one free parameter is enough to control the relative strength of the two components, thus $\beta$ is excluded in Equation \ref{eq:rw_sim}.}

\begin{equation}
 \phi(\mathbf{v}) \leftarrow  \frac{\alpha \phi ^ \prime(\mathbf{v}) + \sum_{u}  W_{u,v} ~\phi(\mathbf{u})} {\alpha + \sum_{u} W_{u,v}} \label{eq:rw_sim} 
\end{equation}

%We can further simplify it by having only one free parameter $\alpha$ to control the relative strength of the two components:

%\begin{equation}
% \hat{\mathbf{X}}_v \leftarrow  \frac{\alpha \mathbf{X}_v + \sum_{(u,v) \in E}  W_{u,v} \hat{\mathbf{X}}_u}{\alpha + \sum_{(u,v) \in E} W_{u,v}} \label{eq:rw_sim} 
%\end{equation}

We can show a correspondence between this method and a biased random walk over $G$, where at any vertex $\mathbf{v} \in V$, the walk faces two options: \Ni with probability $\alpha_v$, the walk stops and returns the prior vector $ \phi'(\mathbf{v})$; and \Nii with probability $\beta_v$, the walk continues to one of $\mathbf{v}$'s neighbors $\mathbf{u}$ with probability proportional to $W_{u,v}$. Formally, the random walk view has the following iterative update \cite{Talukdar.Koby.09}:

\vspace{-0.1in}
\begin{equation}
  \phi(\mathbf{v}) \leftarrow  \alpha_v  \phi'(\mathbf{v}) + \beta_v \frac{W_{u,v}}{\sum_u W_{u,v}}  \phi(\mathbf{u}) \label{eq:rw-first-order}
 %\frac{ \beta \sum_{(u,v) \in E}   \hat{\mathbf{X}}_u}{\alpha_v +  \sum_{(u,v) \in E} W_{u,v}} 
\end{equation}

\subsubsection{Retrofitting with Node2Vec}
\label{subsec:retro_node2vec}
The above retrofitting method is limited to only \emph{first-order} proximity, i.e., only immediate neighbors are considered in the objective function  (see Equation \ref{eq:ret}). However, preserving only local structures may not be sufficient for many applications \cite{Tang.Qu:15}. For example, sentences, which are not  directly connected but share neighbors, are likely to be similar, thus should have similar vector representations. 

As described in Section \ref{sec:node2vec}, the controlled random walk in Node2Vec gives us more flexibility in exploring the network structure by combining BFS and DFS search strategies adaptively. We therefore leverage the random walk of Node2Vec to generate neighborhood samples  $N(\mathbf{v})$, but modify the objective in Equation \ref{eq:n2v_int} for retrofitting as follows:

\begin{equation}
%J(\phi) &=& \sum_{u \in V} \Big[ \alpha_u \log P(N(u)|\phi(u)) + \beta_u ||\phi(u) - \phi ^ \prime(u)||^2 \Big] \\
%J(\phi) = \hspace{-0.1cm} \sum_{u \in V} \Big[ \alpha_u \hspace{-0.2cm} \sum_{v \in N(u)} \log P(\phi(v)|\phi(u)) + \beta_u ||\phi(u) - \phi ^ \prime(u)||^2 \Big]  
J(\phi) = \hspace{-0.0cm} - \sum_{\mathbf{v} \in V} \Big[ \alpha_v \hspace{-0.2cm} \sum_{\mathbf{n}_i \in N(\mathbf{v})} \hspace{-0.1cm} \log P(\phi(\mathbf{n}_i)|\phi(\mathbf{v})) + \beta_v ||\phi(\mathbf{v}) - \phi ^ \prime(\mathbf{v})||^2 \Big]  \label{n2v-retro}
\end{equation}

The first component of the model minimizes the cross entropy (or KL divergence) with the neighbor distribution, where the second component encourages the induced vectors to be close to their prior values.   
%We can define a variant of the above model that minimizes the cross entropy with respect to the weight distributions of the neighbors.      

%\begin{equation}
%J(\phi) = \hspace{-0.1cm} \sum_{u \in V} \Big[ \alpha_u \hspace{-0.2cm} \sum_{v \in N(u)} \hspace{-0.2cm} w_{u,v} \log P(\phi(v)|\phi(u)) + \beta_u ||\phi(u) - \phi ^ \prime(u)||^2 \Big]  
%\end{equation}

% \begin{itemize}
% \item \red{We want to show an example of a ego-network and how we are updating a particular node representation using this model to visualize the steps.}
% \end{itemize}

\subsection{Dis-S2V by Regularization} \label{sec:reg_sen2vec}
% !TEX root = sen2vec-www-17.tex
%%%
%%% Regularized Sen2Vec models
%%%

Rather than retrofitting the vectors from the content-based model on the discourse graph as a post-processing step, we can incorporate the neighborhood information directly into the objective function of the content-based model as a regularizer, and learn the Dis-S2V vectors in a single step. We define the following objective to minimize:      

\begin{equation}
J(\phi) = \sum_{\mathbf{v} \in V} \Big[ L(\mathbf{v}) + \hspace{-0mm} \beta \hspace{-0.2cm} \sum_{(u,v)\in E} \hspace{-0.5mm}  W_{u,v} ||\phi(\mathbf{u}) - \phi(\mathbf{v})||^2 \Big] \label{eq:reg}
\end{equation} 

\noindent where the first component $L(\mathbf{v})$ refers to the negative log likelihood loss of the content based model described in Section \ref{sec:sen2vec}, i.e., Equation \ref{eq:dm} for DM and Equation \ref{eq:dbow} for DBOW. The second component is the graph smoothing regularizer with $\beta$ being the regularization strength. In our experiments we use the DBOW model as our content-based model. Since this model learns the vectors from scratch in one shot by considering information from both sources, the two components can be better adjusted to produce better quality vectors.

\subsubsection{Incorporating Word Semantics}
\label{subsec:dictreg}
%The Dis-S2V models described so far trains the vectors by exploiting the content of a sentence and its neighborhood in the discourse graph. 

Notice that the content-based models also train word vectors as a byproduct (i.e., $\phi$ and $\omega$). Recent work has shown that lexical semantic relations (e.g., synonymy, hypernymy) encoded in semantic lexicons (e.g., WordNet \cite{Miller:95}, Framenet \cite{Baker:98}) can improve the quality of word vectors that are trained solely on the sentences \cite{Xu:2014,faruqui:2015:Retro,yu:dredze:2014}. Therefore, in addition to the discourse graph, semantic lexicons can be used as a word-level information source to refine the word vectors in the model, which should in turn refine the sentence vectors. Following previous work \cite{Xu:2014,yu:dredze:2014}, we first convert any kind of semantic lexicon into a network, where semantically similar words defined by the lexicon become neighbors. Then we induce this information as another word-level graph smoothing factor in the objective function of the DBOW model.

\subsubsection{Other Possible Extensions}

Notice that the regularizer in Equation \ref{eq:reg} considers only first-order neighborhood. One possible extension would be to consider the neighborhood sampled by the stochastic sampling method of Node2Vec, which is more flexible. Another interesting extension of the model would be, rather than defining the regularizer as a weighted distance between two vectors, we can use the objective of Node2Vec as the regularizer and minimize the combined negative log likelihood.

\section{Experimental Settings}
\label{settings}
% \red{
% - Primary: Naffi  \\
% - Assist: Tanay \& Shafiq
% }
%\subsection{Evaluation Tasks \& Datasets}
%\subsection{Experimental Results}
%In this section, we present our results on three different sentence level tasks: (1) Classification (2) Clustering, and (3) Ranking. Table~\ref{tab:rank-duc}, ~\ref{tab:rank-duc} and ~\ref{tab:rank-duc} show results for these three tasks. We organize our models into $5$ blocks. Block I represents result for the \ptov\ model. Block II represents the variants of the node2vec model (which uses stochastic sampling method to build corpus of sentences). Block III, IV and V show results for weighted and unweighted versions of Iterative Update, Regularized and Dictionary based regularizer, respectively. Note that both of the regularized models incorporate discourse information during training whereas the Iterative Update model retrofits trained representation of sentences using the discourse information captured by the discourse graph.

In this section, we describe our experimental setup. Specifically, we explain the tasks, datasets, metrics and experimental protocol that we use to demonstrate our models' efficacy.

\subsection{Task Description and Datasets}
\label{subsec:tasks_dataset}
We experiment on three different sentence level tasks: \Ni \emph{classification}, \Nii \emph{clustering}, and \Niii \emph{ranking}. These are the three fundamental information system tasks and good performance over the tasks will indicate the robustness of our models in a wide range of applications. For classification (or clustering), we measure the performance of our models in classifying (or grouping) sentences based on their topics, whereas in Ranking, we investigate our model's performance in ranking the most central topical sentences by evaluating it on an \emph{extractive summarization} task \cite{Nenkova:McKeown:11}.

%consider the task of    

%For Classification, our goal is to correctly classify top-$2\%$ of the extracted sentences into their corresponding category. In Clustering, we measure the performance of our models in assigning key sentences from the same topic in the same cluster whereas in Ranking, we investigate our model's performance in ranking the most central topical sentences by evaluating it on extractive summarization task. Extractive summarization task (automatic summarization) involves condensing a document  to produce a human comprehensible summary. It is different from abstractive summary which can rewrite the sentences it returns compared to extractive summary which directly returns one or more sentences from the document. 

%\red{We need to say why three different tasks are required. What are the insights these tasks can provide about our model? Mention dumping factor in page-rank}

\subsubsection{Datasets for Classification and Clustering}

We use \emph{20-Newsgroups} and \emph{Reuters-21578} datasets for the classification and clustering tasks. These datasets are publicly available and widely used in these tasks. 

%and DUC 2001, 2002 datasets for ranking (summarization) task. 

%These datasets are publicly available and widely used in the corresponding tasks. We briefly describe the datasets below.

\begin{table}[t!]
\centering
\ra{1.0}
\resizebox{\columnwidth}{!}{%
\begin{tabular}{l|ccccc}
\toprule
Dataset       	& \# Doc. 		    & \# Sen. (total) & \# Sen. (annot.)	& \# Class 	\\
\midrule
\emph{20 Newsgroups} 	& 7781            	& 95809           	&	22374	&	8           		\\
\emph{Reuters-21578} 	& 9001            	& 42192             & 	13305	&	8           		\\
%DUC 2001		& 486				& 19549				& N/A				\\
%DUC 2002		& 471				& 13129				& N/A				\\
\bottomrule
\end{tabular}%
}
\caption{Basic statistics of the datasets for topic classification and clustering}
\label{table-datasets}
\end{table}

\noindent \textbf{\emph{20 Newsgroups}:} This is a collection of approximately $20,000$ news documents.\footnote{http://qwone.com/~jason/20Newsgroups/} These documents are organized into $20$ different topics. Some of these topics are closely related (e.g., \emph{talk.politics.guns} and \emph{talk.politics.mideast}), while others are diverse in nature (e.g., \emph{misc.forsale} and \emph{soc. religion.christian}). We selected $8$ diverse topics in our experiments from the $20$ topics. The selected topics are: \emph{talk.politics. mideast}, \emph{comp.graphics}, \emph{soc.religion.christian}, \emph{rec. autos}, \emph{sci.space}, \emph{talk.politics.guns}, \emph{rec.sport.baseball}, and \emph{sci.med}. 

\noindent \textbf{\emph{Reuters-21578}:} The is a collection of articles which appeared on the Reuters newswire in 1987.\footnote{http://kdd.ics.uci.edu/databases/reuters21578/} It has $21578$ documents covering $672$ topics. We use ``ModApte'' for training-test split and select documents only from the most $8$ frequent topics. The topics are: \emph{acq, crude, earn, grain, interest, money-fx, ship,} and \emph{trade}. Table ~\ref{table-datasets} shows some basic statistics on the resultant datasets. 

%Table ~\ref{table-datasets} shows basic statistics of this dataset. There are $9001$ documents containing $42192$ sentences in this dataset. 

\noindent \textbf{Generating Sentence-level Topic Annotations:} For our evaluation on sentence-level topic classification and clustering tasks, we have to create topic annotations at the sentence-level from the document-level topic labels. One option is to assume all the sentences of a document to have the same topic label as the document. However, this naive assumption propagates a lot of noises. Although sentences in a document collectively address a common topic,  not all sentences are directly linked to that topic, rather they play supporting roles. To minimize this noise, we use an extractive (unsupervised) summarizer, LexRank \cite{Erkan.Radev:04}, to select the top $P\%$ sentences as representatives of the document and label them with the same topic label as the document; see Sec. \ref{subsec:summary-gen} for details about the LexRank method. In our experiments, we use the S2V vectors from the content-based model (Sec. \ref{sec:sen2vec}) to represent a node (sentence) in LexRank and we set $P=2$. The fourth column in Table \ref{table-datasets} shows the total number of topic annotated sentences in each dataset that we use for classification and clustering evaluation tasks.

\subsubsection{Datasets for Extractive Summarization}

For summarization, we use the benchmark datasets from DUC $2001-02$, where the tasks were to generate a $10$-words summary and a $100$-words summary for each document in the datasets.\footnote{http://www-nlpir.nist.gov/projects/duc/guidelines} Table \ref{table-duc-datasets} shows some basic statistics about the datasets. DUC-$2001$ has $486$ documents whereas DUC-$2002$ has $471$ documents. The average number of sentences per document is $40$ and $28$, respectively. For each document, $2$-$3$ short reference (human authored) summaries are available, which we use as  gold summaries in our evaluation. The human authored summaries are of approximately $100$ words. On average, the datasets have $2.17$ and $2.04$ human summaries per document, respectively.

%This dataset has multiple short summaries (approximately 100 words) for each document. There are total 1063 short summaries; on average about 2.17 for each document.
%The DUC 2002 has 471 documents where each document has about 28 sentences and 2.04 short summaries on average. For both DUC 2001 and 2002,

\begin{table}[t!]
\centering
\ra{1.0}
\resizebox{\columnwidth}{!}{%
\begin{tabular}{l|cccc}
\toprule
Dataset       	& \# Doc. 		    & \# Sen. (Avg) 		& \# Sum. (Avg) 	\\
\midrule
%20 Newsgroups 	& 7781            	& 95809           	& 8           		\\
%Reuters-21578 	& 9001            	& 42192             & 8           		\\
DUC 2001		& 486				& 40				& 2.17				\\
DUC 2002		& 471				& 28				& 2.04				\\
\bottomrule
\end{tabular}%
}
\caption{Basic Statistics of the DUC Datasets}
\label{table-duc-datasets}
\end{table}

\subsection{Discourse Network Statistics}

For constructing our discourse graph, we use the vectors learned from \ptov (Sec. \ref{sec:sen2vec}). Every sentence is eligible to connect with any other sentence in the dataset. However, we restrict the connections to make the graph reasonably sparse for computational efficiency, while still ensuring that the graph is informative enough. We achieve this by imposing two kinds of constraints based on the similarity values. 

First, we restrict the edges by setting thresholds for intra- and across-document connections; sentences in a document are connected only if their cosine similarity is above the intra-document threshold, similarly, sentences across documents are connected only if their similarity is above the across-document threshold. We use $0.5$ and $0.8$ for intra- and  across-document thresholds, respectively. Second, we further prune the network by keeping only the top $20$ similar neighbors for each node. Table~\ref{tab:dis-net} shows the basic statistics of the resultant discourse graphs for all of our datasets.

%define  to ,  

%  \red{Morever, for each sentence we keep top-$20$ neighbors based on the similarity from the pivot sentence.} 

%\red{How is the network built? Avg. degree of nodes. Inter and Intra parameters.}
\begin{table}[t!]
\centering
\begin{tabular}{l|cccc}
\toprule
Dataset       	& \# Nodes 		& \# Edges 		& 	Avg. \# Edges	\\
\midrule
\emph{20 Newsgroups} 	&   95809       &	1370149		&	14.03			\\
\emph{Reuters-21578} 	&   42192       &	471163     	&	11.17			\\
DUC 2001		& 	19549		&	321423		&	20.15			\\
DUC 2002		& 	13129		&	216492    	&	16.49			\\
\bottomrule
%\hline
\end{tabular}
\caption{Discourse Network Statistics. Here to note that, discourse network may have multiple connected component.}
\vspace{-0.2in}
\label{tab:dis-net}
\end{table}

\subsection{The Extractive Summarizer} \label{subsec:summary-gen}

We use the graph-based LexRank algorithm \cite{Erkan.Radev:04} to rank the sentences in a document based on their importance. To get the summary sentences of a document, we first build a weighted graph, where the nodes represent the sentences of the document and the edges represent the cosine similarity between learned representations (using one of the models in Section \ref{methods}) of the two corresponding sentences. To make the graph sparse, we avoid edges with weight less than $0.10$. We run the PageRank algorithm \cite{Page.Brin:99} on the graph to determine the rank of each sentence in the document, and thereby \emph{extract} the key sentences as summary of that document. The dumping factor in PageRank was set to $0.85$.

%  sentence-sentence network with all the sentences in that document. The edge-weights are based on the cosine similarity between learned representations of two sentences. 
  
%   In our case, the threshold is $0.10$. 

%\red{Did we add the page-rank paper in the reference?} 
%To evaluate the summary In our ROUGE configuration, we avoid stop words. We simply provide all the ranked sentences to ROUGE and let  find $10/100$ words to compare with the $10/100$ words in reference summary.

\subsection{Evaluation Metrics} \label{subsec:metrics}

\noindent \textbf{Topic Classification:} We use precision (P), recall (R), accuracy (Acc), F1 measure (F1), and Cohen's Kappa ($\kappa$) as evaluation metrics for the classification task.  

\noindent \textbf{Topic Clustering:} For measuring clustering performance, we use various  measures such as homogeneity score (H), completeness score (C), V-measure ~\cite{Rosenberg.Hirschberg:2007} (V), and adjusted mutual information (AMI) score. The idea of homogeneity is that the class distribution within each cluster should be skewed to a single class. Completeness score determines whether all members of a given class are assigned to the same cluster. The harmonic mean of these two measures is the V-measure. AMI measures the agreement of two assignments, in our case the clustering and the class distribution. It is normalized against chance. All these measures are bounded by $[0, 1]$. Higher score means a better clustering. 

\noindent \textbf{Summarization:} We use the widely used automatic evaluation metric ROUGE \cite{Lin:04} to evaluate the system-generated summaries. ROUGE is a recall oriented metric that computes $n$-gram recall between a candidate summary and a set of reference (human authored) summaries. Among the variants, ROUGE-1 (i.e., $n=1$) has been shown to correlate well with human judgments for short summaries \cite{Lin:04}. Therefore, we only report ROUGE-1 in this paper. The configuration for ROUGE in our case is \emph{-c $99$ -2 -1 -r $1000$ -w $1.2$ -n $4$ -m -s -a -l 10} (or $10$). Depending on the task at hand, ROUGE collects the first $10$ or $100$ words from the summary after removing the stop words to compare with the corresponding reference summaries.      

%-1~ as an evaluation metric. ROUGE-1 is a recall oriented metric. It computes 1-gram  For a particular model generated summary, $ms$ and a reference summary, $rs \in RS$, where $RS$ is the set of reference summaries, ROUGE-1 is defined as follows:

%\begin{equation}
%\text{ROUGE-1} = \frac{\forall_{rs \in RS} Count_{match (rs,ms)}(gram_1)}{\forall_{rs \in RS}Count(gram_1)}
%\end{equation}

%The configuration for ROUGE in our case is \emph{-c $99$ -2 -1 -r $1000$ -w $1.2$ -n $4$ -m -s -a} which indicates that we are using 99 percent confidence interval (c) with no gap length limit (-2). We are using Porter stemmer (-m), removing stop-words(-s) and using 1000 samples (r). The max-ngram (m) is set as 4 and we use the model average scoring formula (default).

%%%%%%In case of classification or clustering tasks, we take the top $2\%$ of the ranked sentences as a summary of a document. For the ranking task, we take all the ranked sentence as the summary and compare that with the reference summaries using ROUGE~\footnote{http://www.berouge.com/Pages/default.aspx}, a software package for automated evaluation of summaries. 

\subsection{Experiment Protocols} \label{subsec:exp_protocol}

\begin{table}[t]
\centering
\ra{0.6}
\resizebox{\columnwidth}{!}{%
\begin{tabular}{p{3cm}|p{3cm}|p{5cm}}
\toprule 
Models (Ref) & Variants & Comment \\\midrule
\ptov\ (Sec.~\ref{sec:sen2vec})  & N/A  & Content-based model (baseline). \\\midrule
\ntov\ (Sec.~\ref{sec:node2vec}) & \ntov, \ntovi\ & Network-based model. \ntovi\ initializes the vectors with the vectors learned from \ptov. \\\midrule
Dis-S2V by Retrofitting (Sec.~\ref{sec:retro_sen2vec}) & \itw, \ituw, \ntovr\  (Sec.~\ref{subsec:retro_node2vec}).  & First two models use iterative update and the third one uses regularized N2V objective (Eq. \ref{n2v-retro}). Subscript `w'/`uw' means weighted/ unweighted discourse graph.   \\\midrule
Dis-S2V by Regularization (Sec.~\ref{sec:reg_sen2vec}) & \regw,  \reguw,  \dictregw, \dictreguw. & The regularizer in the first two models are based on the discourse graph, while the other two also use a regularizer based on the semantic lexicon, WordNet (Sec. \ref{subsec:dictreg}). Subscript `w'/`uw' means weighted/unweighted discourse graph. The graph for semantic lexicon is always unweighted.\\
\bottomrule
\end{tabular}}
\caption{Model Variants}
\label{tab:model_variants}
\end{table}

%are weighted and un-weighted variants of the model described in Section~\ref{sec:reg_sen2vec}. Similarly, the last two models are weighted and un-weighted variant of the model in Section~\ref{subsec:dictreg}. 

\noindent \textbf{Model Variants:}
In Table~\ref{tab:model_variants}, we briefly describe our core models and their variants. For detailed discussion, please see the referred sections.

\begin{table}[tb!]
\centering
\ra{0.6}
\begin{tabular}{l|l|cc}
\toprule
Models     & Comment	& Param. 	&	Exp. Set		\\
\midrule
\ptov 		& Sen2Vec    &  $k$   	&	$[8, 10, 12]$	\\ 
\ntovr		& N2V retrofit  & $\beta$			&	$[0.3, 0.6, 0.8, 1.0]$\\ 
\itu		& Iterative retro. & $\alpha$ 		&	---Same as above---\\  
\reg		& Regularized & $\beta$			 &	---Same as above---\\ 
%\reguw		&	$\beta$			&	---Same as above---\\ 
%\dictregw	&	$\beta$			&	---Same as above---\\ 
%\dictreguw	&	$\beta$			&	---Same as above---\\ 
\bottomrule
%\hline
\end{tabular}
\caption{Model Parameters}
\label{tab:model-param}
\end{table}

\noindent \textbf{Model Settings:} All the models except the (iterative) retrofitted ones (i.e., \itw, \ituw) were trained with stochastic gradient descent (SGD), where the gradient is obtained via backpropagation. We used subsampling of frequent words and negative sampling in the classification layer as described in \cite{Mikolov_distributed:2013}, which give significant speed-ups in training.

Table ~\ref{tab:model-param} shows the hyper-parameters of our models and the set of values we tuned with for these hyper-parameters. For each dataset, we randomly selected $20$\% documents from the whole set to form a held-out validation set on which we tune these parameters. To find the best parameter values, we optimize F1 for classification, AMI for clustering and ROUGE-1 for summarization on the validation set. We also choose the parameters internal to the Node2Vec in a similar fashion. Table ~\ref{table:optimal-parameters} shows the optimal values of each hyper-parameter for the four datasets. We evaluate our models on the test set with these optimal values. We run each test experiment five times and take the average to avoid any random behavior appearing in the results. We observed the standard deviation to be quite low.

%\red{We need to say somewhere the modification we did for node2vec: instead of pre-calculating all the transition probability we use caching to fit in our small memory.}

%Once the optimal values are determined, we use the values in the test set experiment.
 %We find the optimal values for these parameters with respect to the above mentioned metrics.

% Section ~\ref{results} describes our experiment results in detail for each of the tasks mentioned earlier. 

%\red{\textbf{Ranking Task Parameters:} There are $97$ and $94$ documents in the validation set of DUC 2001 and DUC 2002, respectively. For DUC 2001, we find the optimal values of the hyper-parameters as $10$, $1.0$, $1.0$, $0.8$, $1.00$, $0.60$ and $0.30$ for the \ptov, \ntovr, \ituw, \regw, \reguw, \dictregw and \dictreguw models, respectively. For DUC 2002, the corresponding values are $12$, $1.0$, $1.0$, $0.3$, $0.3$, $0.80$ and $0.60$.}

\begin{table}[t!]
\centering
\ra{0.6}
\resizebox{\columnwidth}{!}{%
\begin{tabular}{l|l|l}
\toprule
Dataset	&	Task	      & <$k$,  $\beta$,  $\alpha$, $\beta$, $\beta$, $\beta$, $\beta$>	\\
    \midrule
Reuters-21578 & class     & <12, 0.3, 0.6, 0.8, 0.6, 0.6, 0.6> \\
              & clust     & <12, 0.3, 0.3, 1.0, 1.0, 1.0, 0.6> \\
20 Newsgroups & class     & <10, 0.6, 0.3, 0.6, 0.8, 1.0, 0.6> \\
              & clust     & <10, 0.3, 0.3, 1.0, 1.0, 1.0, 1.0> \\
\midrule
DUC 2001      & ranking       & <10, 1.0, 1.0, 0.8, 1.0, 0.6, 0.3> \\
DUC 2002      & ranking       & <12, 1.0, 1.0, 0.3, 0.3, 0.8, 0.6>\\
\bottomrule
\end{tabular}%
}
\caption{Optimal values of the parameters.}
\label{table:optimal-parameters}
\end{table}

\section{Results}
\label{results} 
% \red{
% - Primary: Tanay \& Naffi  \\
% - Assist: Shafiq
% }

\begin{table*}[!ht]
\centering
\ra{1.2}
\resizebox{\linewidth}{!}{%
\begin{tabular}{@{}cllllllllllll@{}}
\toprule
\multicolumn{2}{l}{}	&	\multicolumn{5}{c}{\emph{Reuters-21578}}         & \phantom{abc}          & \multicolumn{5}{c}{\emph{20 Newsgroups}}                   \\
\multicolumn{2}{l}{}	&	R        & P        & F1       & Acc      & $\kappa$	&    & R        & P        & F1       & Acc      & $\kappa$    \\ 
\cline{3-7}\cline{9-13}
\\
 \multirow{1}{*}{\textbf{I}}	& \ptov	&	84.20    & 84.00    & 83.80    & 84.24    & 79.84    &	& 79.00    & 79.80    & 79.00    & 79.23    & 75.90    \\
\midrule
\midrule
\multirow{2}{*}{\textbf{II}}	& \ntov	&	($-$) 1.40 & ($-$) 1.60 & ($-$) 1.40 & ($-$) 1.50 & ($-$) 1.94 &	& ($-$) 2.00 & ($-$) 2.80 & ($-$) 2.00 & ($-$) 2.15 & ($-$) 2.46 \\
&	\ntovi	&	($-$) 0.20 & ($-$) 0.20 & (+) 0.00     & ($-$) 0.20 & ($-$) 0.18 &	& ($-$) 1.00 & ($-$) 1.80 & ($-$) 1.00 & ($-$) 1.49 & ($-$) 1.71 \\\midrule
\multirow{3}{*}{\textbf{III}} &	\ntovr	&	(+) 2.60 & (+) 2.80 & (+) 2.40 & (+) 2.52 & (+) 3.24 &	& (+) 3.00 & (+) 2.60 & (+) 3.00 & (+) 2.98 & (+) 3.46 \\

	& \ituw	&	(+) 2.40 & (+) 2.80 & (+) 1.80 & (+) 2.20 & (+) 2.80 &	& (+) 2.00 & (+) 2.00 & (+) 2.00 & (+) 2.08 & (+) 2.40 \\
&	\itw	&	(+) 2.00 & (+) 2.60 & (+) 1.40 & (+) 2.00 & (+) 2.50 &	& (+) 2.00 & (+) 1.40 & (+) 2.00 & (+) 1.97 & (+) 2.26 \\
\midrule
\multirow{4}{*}{\textbf{IV}}	& \reguw	&	\textbf{(+) 3.60} & \textbf{(+) 3.80} & \textbf{(+) 3.40} & \textbf{(+) 3.42} & (+) \textbf{4.32} &	& (+) 2.00 & (+) 2.20 & (+) 2.20 & (+) 2.19 & (+) 2.55 \\
&	\regw	&	(+) 3.40 & (+) 3.60 & \textbf{(+) 3.40} & (+) 3.30 & (+) 4.22 &	& \textbf{(+) 3.40} & \textbf{(+) 3.20} & \textbf{(+) 3.60} & \textbf{(+) 3.26} & \textbf{(+) 3.79} \\
& \dictreguw	&	(+) 3.20 & (+) 3.20 & \textbf{(+) 3.40} & (+) 3.26 & (+) 4.18 &	& (+) 3.00 & (+) 2.20 & (+) 2.80 & (+) 2.38 & (+) 2.76 \\
&	\dictregw	&	(+) 2.80 & (+) 2.80 & (+) 2.80 & (+) 2.60 & (+) 3.32 &	& (+) 2.60 & (+) 2.20 & (+) 2.40 & (+) 2.32 & (+) 2.69	\\
\bottomrule
\end{tabular}%
}
\caption{Performance of our models in \emph{Reuters-21578} and \emph{20 Newsgroups} datasets in the classification task. The results are averaged over $5$ different runs. In Block II, III, and IV, we show our model's performance in comparison to the baseline, \ptov.}
%for Recall (R), Precision (P), Accuracy (Acc) and Cohen' Kappa ($\kappa$).}
\label{tab:class}
\end{table*}

%In this section, we present our results in three sentence level tasks: (1) Classification, (2) Clustering and (3) Ranking as described in section~\ref{subsec:tasks_dataset}. 

Tables~\ref{tab:class},~\ref{tab:clustering} and~\ref{tab:rank-duc} present the results on topic classification, topic clustering and summarization tasks, respectively. In the tables, we organize our models into $4$ blocks; see also Table \ref{tab:model_variants} for a brief description of these blocks.  

Block I shows results for the baseline \ptov\ model described in Sec.~\ref{sec:sen2vec}, which is purely a content-based model. \ptov\ model concatenates the vectors learned by DM and DBOW models. The concatenated model performed better than individual ones. We show all other models' relative performance with respect to this \ptov\ baseline.

Block II presents the results for \ntov\ and \ntovi\ described in Sec.~\ref{sec:node2vec}. \ntovi\ initializes the vectors with the representation learned from the content-based model, \ptov. In Block III, we present the results of ``Dis-S2V by Retrofitting'' models described in Sec.~\ref{sec:retro_sen2vec}. The models in this category are: \ntovr, \ituw, and \itw. \ntovr\ uses a regularizer in the original N2V objective, which restricts new vector representation not to go far away from the already learned content-based representation. \ituw\ and \itw\ retrofit the vectors in such a way that the vectors become closer to the one-hop neighborhood in the discourse network. Finally, in Block IV, we show results for ``Dis-S2V by Regularization'' models described in Sec.~\ref{sec:reg_sen2vec}. These models induce information from the discourse network during training, \dictreguw\ and \dictregw\ additionally use the lexicon, WordNet.

%Both \ntov\ and \ntovi\ use stochastic sampling technique to generate context. 

%We present our performance metrics and experimental protocol for each of the tasks in Section~\ref{subsec:metrics} and~\ref{subsec:exp_protocol}, respectively. The result presented in Block I (first row) are compiled using \ptov\ model, which is our baseline. \ptov\ model concatenates the vectors learned by DM and DBOW models. We show all other models' relative performance with respect to the baseline. %The larger the positive improvement, the better the model is. 

%We describe the performance metrics for each of these tasks in Section~\ref{subsec:metrics}. 
% presents result for classification, Table~\ref{tab:clustering} shows result for clustering and finally, in Table~\ref{tab:rank-duc}, we show our model's performance 
% in ranking task. % : (1) Classification (2) Clustering, and (3) Ranking. Table~\ref{tab:classification}, ~\ref{tab:clustering} and ~\ref{tab:rank-duc} show results for these three tasks. 
% for the variants of the node2vec model (which uses stochastic sampling method to build corpus of sentences). 
% Block III, IV and V show results for weighted and un-weighted versions of Iterative Update, Regularized and Dictionary based regularizer model respectively. Please remember that both of the regularized models incorporate discourse information during training whereas the Iterative Update model retrofits trained representation of sentences using the discourse information captured by the discourse graph.

\subsection{Classification Results} 

The results in Table~\ref{tab:class} demonstrate significant performance improvement of our models in $8$-class \emph{classification} task compared to the baseline, \ptov. For \emph{Reuters-21578} dataset, \reguw\ from Block IV becomes the best performer and it improves over the baseline in R, P, Acc and $\kappa$ metric by $3.60\%$, $3.80\%$, $3.40\%$, $3.42\%$ and $4.32\%$, respectively. The inter-rater metric ($\kappa$) improves over $4\%$, which is a good indicator that \reguw\ has better agreement with the gold label assigner. For \emph{20 Newsgroups}, \regw\ becomes the best performer across all the metrics. A closer observation will reveal that \regw\ also performs competitively with \reguw\ in \emph{Reuters-21578} dataset. This is intuitive as these models use information from both the content and the discourse network.  Moreover, as expected, \ntov\ model performs poorly as it only uses information from the content network. \ntovi\ performs better than \ntov\ as it incorporates the content vector during initialization. \ntovr, \ituw, and \itw\  perform better than \ptov\ as all of these models control the information induction into the new vector in the form of \Ni being closer to the representation learned from the content-based model, or \Nii being closer to the neighbors in the discourse graph. Overall, the models which can effectively use the content and information in discourse network wins over models, which only use content or just the network. 

% ,  performance of our models in the classification task for the  and \emph{20 Newsgroups} datasets. With respect to the baseline, \ptov, our models perform significantly better in terms of the classification metrics (explained in ~\ref{subsec:metrics}) in general. For each dataset, the best performing model for each performance measure is highlighted. We observe that, like in the ranking task, the models in Block IV and V perform better compared to the other models and this is consistent across the datasets. 

\begin{table*}[!ht]
\centering
\ra{1.2}
\resizebox{\linewidth}{!}{%
\begin{tabular}{@{}cllllllllll@{}}
\toprule
\multicolumn{2}{l}{}	&	\multicolumn{4}{c}{\emph{Reuters-21578}}         & \phantom{abc}          & \multicolumn{4}{c}{\emph{20-Newsgroups}}                   \\
\multicolumn{2}{l}{}	&	H        & C        & V       & AMI	&    & H        & C        & V       & AMI    \\ 
\cline{3-6}\cline{8-11}
\\
\multirow{1}{*}{\textbf{I}}	& \ptov	&	45.54     & 39.88     & 42.50     & 39.50     &  & 35.08     & 36.26     & 35.68     & 35.08     \\
\midrule
\midrule
\multirow{2}{*}{\textbf{II}}	& \ntov	&	(+) 8.70  & (+) 7.98  & (+) 8.36  & (+) 8.90  &  & (+) 12.80 & (+) 12.84 & (+) 12.80 & (+) 12.80 \\
&	\ntovi	&	\textbf{(+) 13.56} & \textbf{(+) 12.50} & \textbf{(+) 13.02} & \textbf{(+) 12.80} &  & \textbf{(+) 13.16} & \textbf{(+) 13.22} & \textbf{(+) 13.18} & \textbf{(+) 13.10} \\
\midrule
\multirow{3}{*}{\textbf{III}} &	\ntovr	&	(+) 2.30  & (+) 1.86  & (+) 2.10  & (+) 1.98  &  & (+) 4.06  & (+) 3.60  & (+) 3.84  & (+) 4.04  \\

	& \ituw	&	(+) 3.52  & (+) 2.52  & (+) 2.96  & (+) 2.88  &  & (+) 5.14  & (+) 4.06  & (+) 4.58  & (+) 5.10  \\
&	\itw	&	(+) 3.62  & (+) 2.70  & (+) 3.14  & (+) 2.93  &  & (+) 5.08  & (+) 4.08  & (+) 4.58  & (+) 5.06  \\
\midrule
\multirow{4}{*}{\textbf{IV}}	& \reguw	&	(+) 6.78  & (+) 5.74  & (+) 6.26  & (+) 5.88  &  & (+) 10.98 & (+) 10.80 & (+) 10.90 & (+) 10.96 \\
&	\regw	&	(+) 7.32  & (+) 6.32  & (+) 6.80  & (+) 6.48  &  & (+) 11.40 & (+) 12.00 & (+) 11.68 & (+) 11.36 \\
& \dictreguw	&	(+) 7.18  & (+) 6.62  & (+) 6.92  & (+) 6.90  &  & (+) 10.80 & (+) 10.06 & (+) 10.42 & (+) 10.76 \\
&	\dictreguw	&	(+) 6.58  & (+) 5.76  & (+) 6.18  & (+) 5.93  &  & (+) 11.20 & (+) 10.58 & (+) 10.88 & (+) 11.20 \\
\bottomrule
\end{tabular}%
}
\caption{Performance of our models in \emph{Reuters-21578} and \emph{20 Newsgroups} datasets in the clustering task.}
%The results are averaged over $5$ different runs. In Block II, III, and IV, we show our model's performance relative to the baseline, \ptov\ for metrics: Homogeneity (H), Completeness (C), V-measure (V) and Adjusted Mutual Information (AMI).}
\label{tab:clustering}
\end{table*}

\subsection{Clustering Results}

The \emph{clustering} results in Table~\ref{tab:clustering} clearly shows that all our models outperform the baseline by a good margin, and the models from Block IV perform significantly better than the models in Block III in all the metrics. The adjusted mutual information score (AMI) is greater than $0.40$ for all of our models, indicating that our models are better than the random cluster assignments. One important observation is that \ntovi\ becomes the best performer with a relative improvement around $13\%$ compared to the baseline. Note that, \ntovi\ incorporates information from content-only representation in initializing its model, then it gets updated based on the network neighborhood. \ntov, which uses only the network neighborhood, also performs better than the content-only baseline. This indicates that the neighborhood information is quite beneficial for clustering tasks. These promising results one more time indicate the efficacy of our representation model over the content-only models.

\begin{figure}[t!]
\centering
\includegraphics[width=\columnwidth]{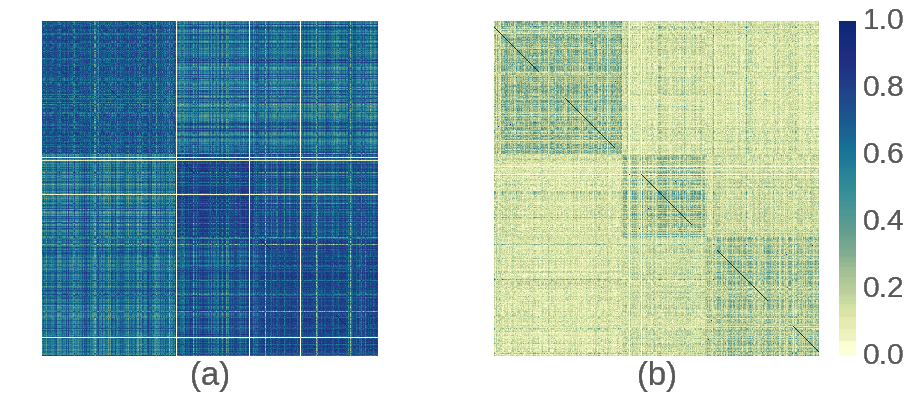}
\vspace{-0.3in}
\caption{Heat-map showing pair-wise similarities between the learned representations from (a) \ptov\ and (b) \ntovi\ for the \emph{20 Newsgroups} dataset.}
\label{fig:heatmap_cluster}
\end{figure}

Figure~\ref{fig:heatmap_cluster} shows a visualization of clustering generated using \ptov\ and \ntovi\ models for only three ($3$) categories out of the $8$ categories. From the figure, it is clear that \ntovi\ was able to retrieve all three clusters whereas \ptov\ is merging two out of three clusters to form one big cluster.

\subsection{Ranking Results}
From Table~\ref{tab:rank-duc}, we observe that over all the summarization tasks, \ntovr, \reguw, \regw, \dictreguw, and \dictregw\ consistently outshine the baseline \ptov\ model. For DUC 2001, in $10$-length summary, \ntovr\ wins by $2.21\%$ over \ptov, whereas for $100$-length summary, \ntovi\ wins over \ptov\ by a margin of $5.16\%$. For DUC 2002, \regw\ and \reguw\ perform better than all other models. For length $10$, \regw\ improves over \ptov\ by $2.5\%$, whereas \reguw\ gets around $3.26\%$ improvement for length $100$. 

In summary, models in Block IV and \ntovr, i.e. models which harness the information content of discourse network during the learning phase have better performance than the other models in the summarization task. Here to note that, \ntovr\ also retrofits during learning as opposed to \itw\ and \ituw, which retrofit the learned representations as a post-processing step.

\begin{table}[!th]
\centering
\ra{1.1}
\resizebox{\linewidth}{!}{%
\begin{tabular}{@{}lcllllll@{}}
\toprule
 & \multicolumn{1}{l}{}                  &                          & \multicolumn{2}{c}{2001}       &    \phantom{abc}& \multicolumn{2}{c}{2002}                         \\
 & \multicolumn{1}{l}{}                  &                          & \multicolumn{1}{c}{10} & \multicolumn{1}{c}{100}& & \multicolumn{1}{c}{10} & \multicolumn{1}{c}{100} \\
\midrule
 & \multirow{1}{*}{\textbf{I}}                  & \ptov                  & 12.65     & 43.63     & & 14.39     & 54.51     \\\midrule \midrule 
 & \multirow{2}{*}{\textbf{II}} & \ntov                      & (+) 1.26        & (+) 5.00   &      & ($-$) 0.45        & ($-$) 0.39    \\
 &                                       & \ntovi               & (+) 0.48        & \textbf{(+) 5.16}   &      & ($-$) 0.87        & ($-$) 0.45    \\ \midrule
  &   \multirow{3}{*}{\textbf{III}}                                      & \ntovr        & \textbf{(+) 2.21}        & (+) 1.88   &      & (+) 0.26        & (+) 0.31    \\
 &                                    & \ituw        & ($-$) 0.20        & ($-$) 1.51   &      & (+) 0.34        & ($-$) 0.17    \\
 &                                       & \itw          & (+) 0.04        & ($-$) 1.02    &     & (+) 0.29        & ($-$) 1.05    \\\midrule
 &   \multirow{4}{*}{\textbf{IV}}                                    & \reguw            & (+) 1.02        & (+) 1.31   &      & (+) 2.44        & \textbf{(+) 3.26}    \\
 &                                       & \regw               & (+) 0.74        & (+) 1.07   &      & \textbf{(+) 2.51}        & (+) 3.10    \\
 &                                    & \dictreguw      & (+) 1.79        & (+) 1.33   &      & (+) 2.04        & (+) 2.71    \\
 &                                       & \dictregw        & (+) 1.46        & (+) 1.79   &      & (+) 1.63        & (+) 2.51    \\ 

\bottomrule
\end{tabular}
}
\caption{Summarization result on DUC 2001-02 datasets for summary lengths of 10 and 100. Here, we present the increment in average recall of a model with reference to the \ptov\ model.} %A model is better if it has larger increment than the other model.}
\label{tab:rank-duc}
\end{table}

In conclusion, over all the three (3) fundamental information system tasks, models in Block IV consistently improve over the \ptov\ model. The models which use both the content and the information in the discourse network win over the baseline in both the classification and the clustering tasks (Please see the results of Block III and IV in Tables~\ref{tab:class} and~\ref{tab:clustering}).

\section{Related Work}
\label{relwork}
% \red{
% - Primary: Tanay  \\
% - Assist: Shafiq \& Naffi
% }\\

Our work is closely related to two different branches of representational learning research: \Ni distributed representation of textual units (e.g., words, paragraphs), and (2) latent representation of nodes in a network. 

%Both of the areas cast the feature extraction process as a representation learning problem

%Contexts in texts usually  

%For texts, context of a unit directly comes from the adjacent words (content) of a particular sentence.

%In the first line of research, while learning representation,  

%For the second kind, contexts are generated via random-walk over a network (in our case discourse network), i.e. from the neighborhood structure of a node. Both of the areas cast the hand-engineered feature extraction process as a representation learning problem i.e. their core philosophy is to automate the feature extraction process while keeping the performance of the task intact or by improving it.

%~\cite{Bengio.Courville:13}

\subsection{Representation for Textual Units}

The Word2Vec model~\cite{Mikolov.Chen:13} to learn distributed representation of words is very popular for text processing tasks. The model also scales well in practice due to its simple architecture.  ~\cite{Le.Mikolov:14} extended Word2Vec to learn the representation for sentences and documents. The model maps each sentence to an unique id and learns the representation for the sentence using the contexts of words in the sentence -- either by predicting the whole context or by predicting a  word in the context given the rest. In our work, we extend this model to incorporate inter-sentence relations in the form of a discourse graph. We do this using a graph-smoothing regularizer in the original objective function or by retrofitting the original vectors on the discourse graph.

Recently, \cite{Xu:2014,faruqui:2015:Retro,yu:dredze:2014} propose methods to incorporate semantic knowledge into the word representation models. Our overall idea of using discourse as an extra source of information is reminiscent of these studies with  a number of key differences: \Ni the semantic network is given in their case, whereas we construct the network using similarities between nodes; \Nii we explore a purely network-based model for learning sentence representations and combine it with the content-based model, whereas they use the network just as an extra source of information in their content-based model. The network-based model gives us the flexibility to adaptively explore the network neighborhood. 

Skip-Thought~\cite{Kiros.Zhu:15} and Fast-Sent~\cite{Hill.Cho:16} are the most recently proposed sentence representation methods.  Skip-Thought uses an encoder-decoder approach to reconstruct neighboring sentences of an input sentence, whereas Fast-Sent predicts words from the neighboring sentences for learning representation of the current sentence. Our discourse network is more general and can connect any pair of sentences. Skip-Thought is computationally very expensive~\cite{Hill.Cho:16}. Our approach is fundamentally different from these models and gives a light-weight solution to incorporate discourse.  

%Our work  from these models.

%However, 
 %We use discourse network, which can go beyond the paragraph or document boundary. 

% . Four of our models, \regw, \reguw, \dictregw, and \dictreguw, are extensions of this work where we incorporate external information in the form of discourse network. \regw,  and \reguw\ use only the discourse network, \dictregw, and \dictreguw\ use external information provided as a dictionary (e.g. WordNet~\cite{Miller:95}) in addition to the discourse network. 

% also have been proposed recently for learning sentence representation. Skip-thought shows that rich sentence semantics can be inferred from the content of adjacent sentences, however, it is very slow to train. however, our work is fundamentally different from their work from the modeling perspective, i.e., we want to test the hypothesis that, whether inducing information using discourse network improves the three fundamental information system tasks or not. We hope to compare experimentally with these methods in future.  

\subsection{Representation for Nodes in Networks}
DeepWalk~\cite{Perozzi.Al-Rfou:14},  LINE~\cite{Tang.Qu:15} and node2vec~\cite{Grover.Leskovec:16} are some of the recent advancements on learning representation for nodes in a network, and are closely related to our work. These studies show promising results over linear and non-linear matrix factorization techniques, such as PCA~\cite{Jolliffe:02}, Iso-Map~\cite{Tanenbaum.De:00}, and LLE~\cite{Roweis.Saul:00}. They are also related to each other in a sense that they all use discriminative models to learn the representation. These methods differ in how they generate the neighborhood (i.e., context) of a node from the graph. DeepWalk generates contexts using truncated random-walk.  Both LINE and node2vec use alias sampling for generating contexts. node2vec uses a controlled random walk to adaptively combine breadth first search and depth first search. In our work, we use node2vec as the network-based model, and we combine it with the content-based model. 

%Three of our models: \ntov, \ntovi, and \ntovr\ are based on node2vec. 

%However, in our models, we can use any kind of node representation model given that the discourse network is available. 

%However,  designs a control mechanism to balance between exploration and exploitation during the walk over the graph (content generation process). 

% internally use the word2vec~\cite{Mikolov.Sutskever:13}, skip-gram, and negative sampling technique and only differ in how they  content to feed in the word2vec model. 

%Unlike the earlier set of works, here, the context information is not directly available (no content). 

% Even though our models are very similar to some of the existing works, we are the first to integrate the context and the network to learn the representation of a sentence which we think is a novel contribution.

\section{Conclusion and Future Work}
\label{conclusion} 
In this paper, we have proposed a set of novel models for learning vector representation of sentences that consider not only content of a sentence but relations among sentences. The relations among sentences are encoded in a discourse graph, which is then used to build discourse informed sentence representation models. We have explored two different ways to incorporate discourse information: \Ni by retrofitting the vectors learned from a content-based model on the discourse graph, and \Nii by regularizing the content-based model with a graph smoothing factor. 

We evaluated our models on three tasks of classification, clustering, and ranking. Our results show that our models outperform the content-only baseline, and the regularized models perform well in all three tasks whereas the retrofitting models are good for only \emph{classification} and \emph{clustering} tasks.

In future, we would like to extend our models to represent groups in social networks and evaluate on prediction tasks involving social groups. The main difference between a network of sentences and a network of social groups is that groups are sets (order is irrelevant), whereas sentences are sequences. However, considering that our content-based model (DBOW) does not take order into account, we should be able to directly apply our models to social groups.    

%evaluate the performance of our models  in social network.  

%and show that our models perform consistently better than the existing models. This is due to the fact that our models can use efficiently both the content of the sentence and topical structure captured by discourse network. We plan to investigate further insights of our models and also to explore the distributed graph database management systems for handling large discourse networks. 

%ACKNOWLEDGMENTS are optional
%\section{Acknowledgments}
%\input{ack}

\clearpage
%\balancecolumns
\bibliographystyle{abbrv}
\bibliography{sen-vec.bib}

%\appendix
%\input{app}
\end{document}